\documentclass[letterpaper]{article} 
\usepackage{aaai23}  
\usepackage{times}  
\usepackage{helvet}  
\usepackage{courier}  
\usepackage[hyphens]{url}  
\usepackage{graphicx} 
\usepackage{natbib}  
\usepackage{caption} 
\usepackage[utf8]{inputenc} 
\usepackage[T1]{fontenc}    
\usepackage{url}            
\usepackage{booktabs}       
\usepackage{amsfonts}       
\usepackage{nicefrac}       
\usepackage{microtype}      
\usepackage{xcolor}         
\usepackage{multirow}
\usepackage{subcaption}
\usepackage{graphicx}
\usepackage{amsmath}
\usepackage{algorithmicx}
\usepackage{algorithm}
\usepackage{algpseudocode}

\def\onedot{.}
\def\eg{\emph{e.g}\onedot} 
\def\ie{\emph{i.e}\onedot}

\newcommand{\mymin}{\operatornamewithlimits{min}}
\newcommand{\mymax}{\operatornamewithlimits{max}}

\usepackage{xcolor}

\usepackage{newfloat}
\usepackage{listings}
\DeclareCaptionStyle{ruled}{labelfont=normalfont,labelsep=colon,strut=off} 
\floatstyle{ruled}
\newfloat{listing}{tb}{lst}{}
\floatname{listing}{Listing}
\title{State-Conditioned Adversarial Subgoal Generation}
\author {
	Vivienne Huiling Wang\textsuperscript{\rm 1,\rm 2}, 
	Joni Pajarinen\textsuperscript{\rm 2},
	Tinghuai Wang\textsuperscript{\rm 3},
	Joni-Kristian K\"{a}m\"{a}r\"{a}inen\textsuperscript{\rm 1}
}
\affiliations {
	\textsuperscript{\rm 1} Computing Sciences, Tampere University, Finland\\
	\textsuperscript{\rm 2} Department of Electrical Engineering
	and Automation, Aalto University, Finland\\
	\textsuperscript{\rm 3} Huawei Helsinki Research Center, Finland\\
	\{huiling.wang,joni.kamarainen\}@tuni.fi, joni.pajarinen@aalto.fi, tinghuaiwang@huawei.com
}

\usepackage{bibentry}

\begin{document}
	
	\maketitle
	
	\begin{abstract}
		Hierarchical reinforcement learning (HRL) proposes to solve difficult tasks by performing decision-making and control at successively higher levels of temporal abstraction. However, off-policy HRL often suffers from the problem of a non-stationary high-level policy since the low-level policy is constantly changing. In this paper, we propose a novel HRL approach for mitigating the non-stationarity by adversarially enforcing the high-level policy to generate subgoals compatible with the current instantiation of the low-level policy. In practice, the adversarial learning is implemented by training a simple state conditioned discriminator network concurrently with the high-level policy which determines the compatibility level of subgoals. Comparison to state-of-the-art algorithms shows that our approach improves both learning efficiency and performance in challenging continuous control tasks.
	\end{abstract}
	
	\section{Introduction} \label{intro}
	
	Hierarchical reinforcement learning (HRL), in which hierarchical policies learn to perform decision-making at successively higher levels of temporal and behavioral abstraction, has long held the promise to tackle complex  problems with long-term credit assignment and sparse rewards. Among the prevailing HRL paradigms, the goal-conditioned HRL frameworks \cite{DayanH92,schmidhuber1993planning,kulkarni2016hierarchical,vezhnevets2017feudal,NachumGLL18,LevyKPS19,ZhangG0H020,li2020learning} have achieved remarkable success. In goal-conditioned HRL, a high-level policy breaks the original task into a series of subgoals that a low-level policy is incentivized to reach. The effectiveness and efficiency of goal-conditioned HRL relies on reasonable and semantically meaningful subgoals providing a strong supervision signal to the low-level policy. 
	
	Nonetheless, off-policy training of a hierarchy of policies remains a key challenge due to the non-stationary state transitions induced by the hierarchical structure. Specifically, the same high-level action taken under the same state in the past may result in significantly different low-level state transitions due to the constantly changing low-level policy which renders the experience invalid for training. When all policies within the hierarchy are trained simultaneously, the high-level transition will constantly change as long as the low-level policy continues to be updated. 
	However, learning hierarchical policies in parallel is still feasible as long as the high-level policy is able to efficiently adapt itself to the updated versions of low-level policy, and the hierarchical policy stabilizes once the low-level policy has converged to an optimal or near optimal policy. 
	HIRO~\cite{NachumGLL18} and HAC~\cite{LevyKPS19} have made attempts to address this problem by \emph{relabeling} an experience in the past with a high-level action, \ie,~subgoal, to maximize the probability of the past lower-level actions. In other words, the high-level action which induced a low-level behavior in the past may be \emph{relabeled} to a subgoal which is likely to induce the same low-level behavior with the current low-level policy.  However, the relabeling approach does not facilitate efficient training of the high-level policy to comply with the update of low-level policy, which consistently generates incompatible subgoals and deteriorates the non-stationarity issue. Such unfit state transitions in off-policy training lead to improper learning of the high-level value function, therefore negatively affecting high-level policy exploration.

	\begin{figure*}
		\centering
		\begin{subfigure}{.36\textwidth}
			\centering
			\includegraphics[width=.99\linewidth,height=5.8cm]{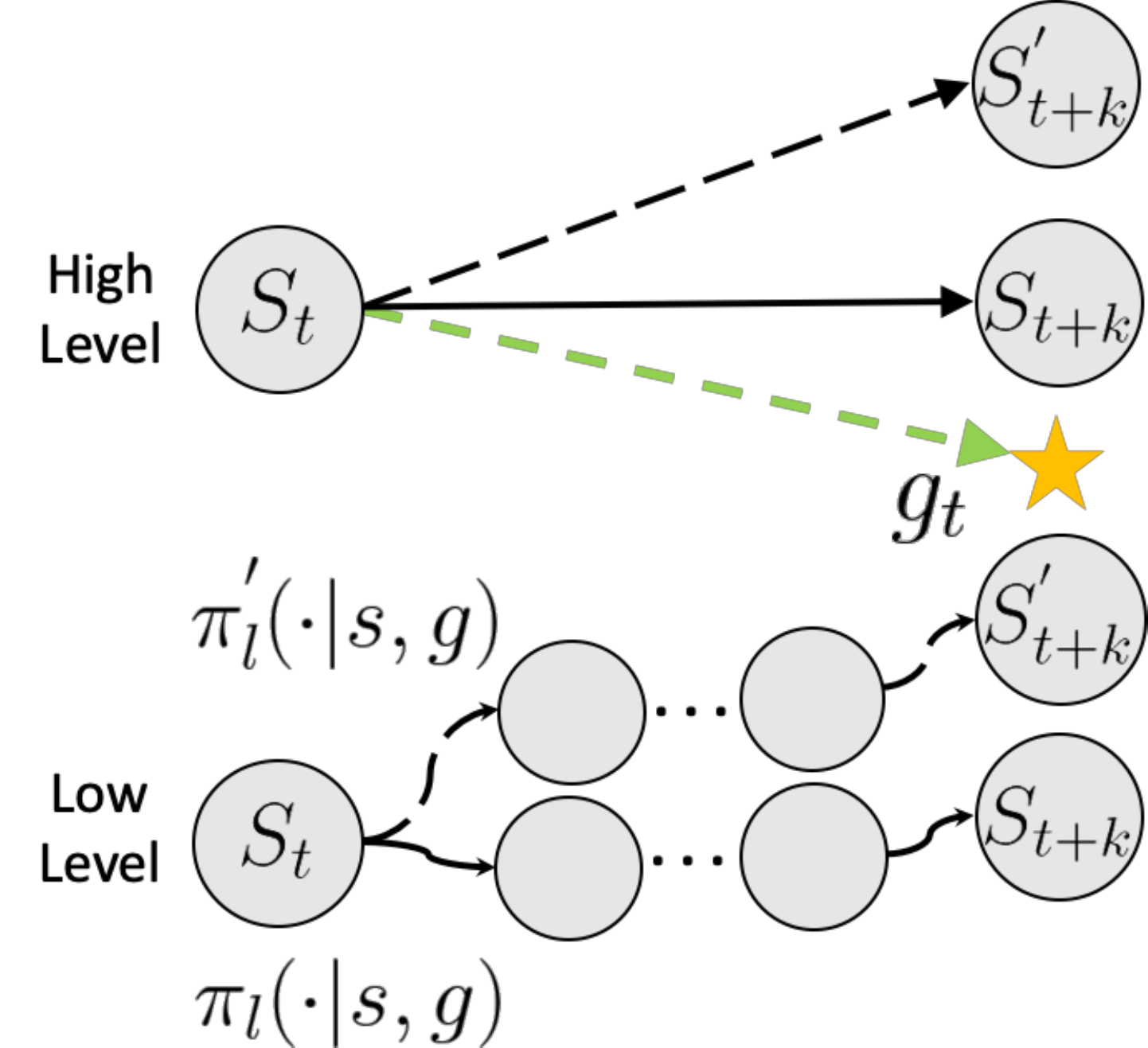}
			\caption{The non-stationarity problem}
			\label{fig:sub1}
		\end{subfigure}%
		\hspace{0.02\textwidth}
		\begin{subfigure}{.61\textwidth}
			\centering
			\includegraphics[width=.99\linewidth,height=5.8cm]{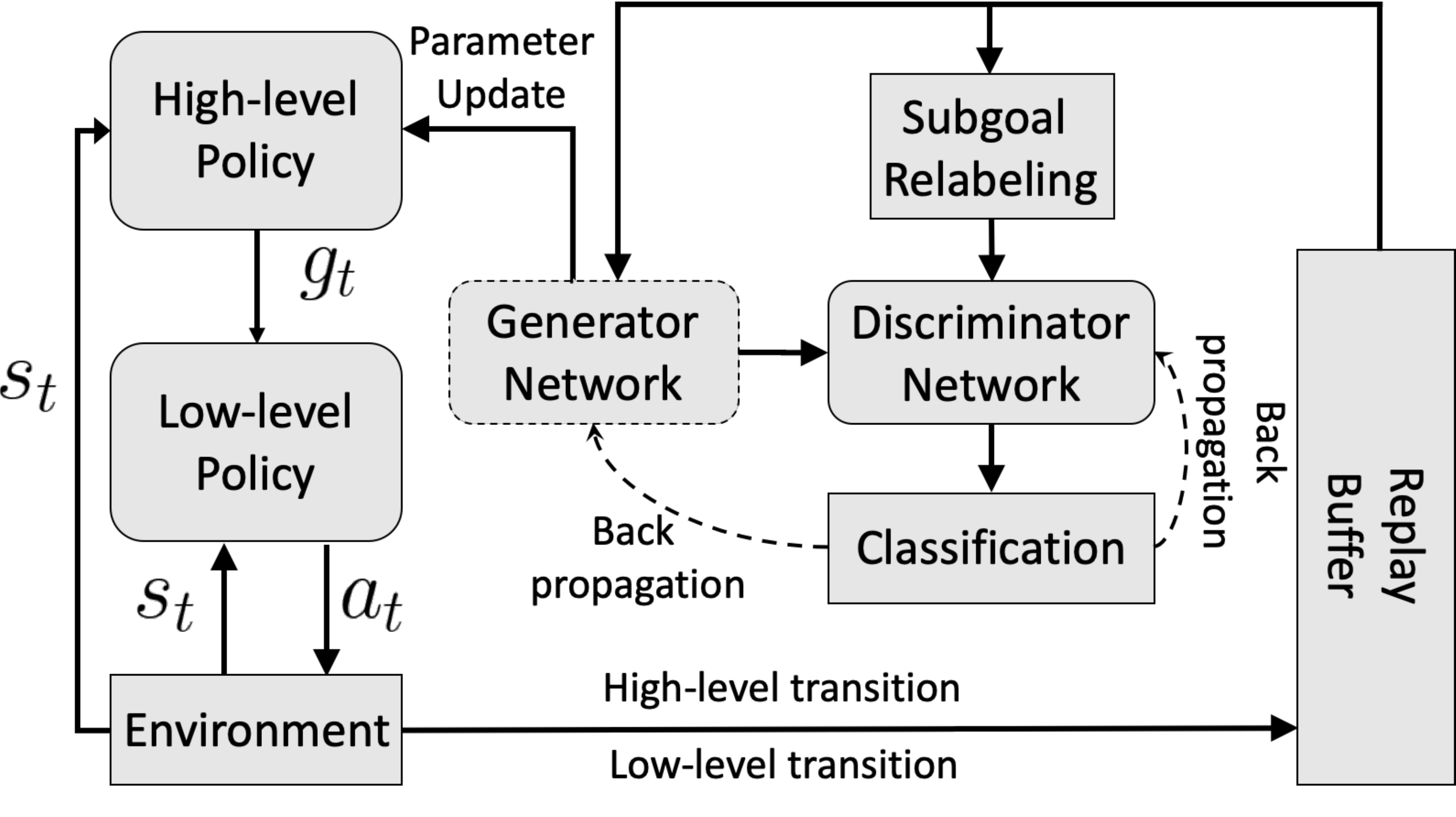}
			\caption{Overview of SAGA}
			\label{fig:sub2}
		\end{subfigure}
		\caption{(a) The emergence of the non-stationarity problem: since the low-level policy has changed from $\pi_l(\cdot|s,g)$ to $\pi^{'}_l(\cdot|s,g)$, a subgoal $g_t$ generated by a certain high-level policy in the past may not yield the same low-level behavior
			given the current low-level policy and consequently renders the experience invalid for training. (b) Overview of SAGA: the high-level policy generates high-level actions \ie, subgoals every $k$ time steps to guide the low-level policy which interacts with the environment. Off-policy adversarial learning is performed for high-level policy to improve its stability and sample efficiency with relabeled subgoals. }
		\label{fig:test}
	\end{figure*}

	In this paper, we present a novel approach for mitigating the non-stationarity in goal-conditioned HRL. We aim to significantly improve the high-level policy’s knowledge of the low-level’s ability, thus improving the overall learning efficiency and stability. 
	Concretely, we introduce an adversarial learning paradigm for HRL which enforces the high-level policy to learn to generate subgoals compatible with the current instantiation of the low-level policy. 
	This is motivated by the assumption that the relabeled subgoals are sampled from a distribution which is asymptotically approximating an optimal high-level policy under stationary data distribution. Consequently the increasing divergence between the distribution of current subgoals and relabeled subgoals is the key indication of the non-stationarity. This suggests a conjecture that once this distribution divergence is mitigated the high-level policy naturally achieves stationarity. 
	
	To this end, we propose a discriminator network to distinguish a generated subgoal that may not be 
	reachable by the low level policy
	from a relabeled subgoal
	that we know is reachable by the low level policy.
	The high-level policy plays the role of the generator network that learns to generate subgoals following a distribution compatible with the current low level policy.
	
	The proposed adversarial learning thus reduces the shift and consequently the divergence in data distribution from relabeled experience to the current high-level policy behaviour and encourages the high level policy to generate reasonable subgoals. Fitting to state transitions with compatible high-level actions effectively improves the accuracy of the high-level value function and enhances its subsequent exploration underpinning a stationary hierarchical model.

	\section{Preliminaries}
	In reinforcement learning, the interaction between agent and environment is modeled as a Markov Decision Process (MDP) $M=<\mathcal{S},  \mathcal{A}, \mathcal{P}, \mathcal{R}, \gamma>$, where $\mathcal{S}$ is a state space, 
	$\mathcal{A}$ is an action set, $\mathcal{P}: \mathcal{S} \times \mathcal{A} \times \mathcal{S} \to [0, 1]$ is a state transition function, 	$\mathcal{R}: \mathcal{S} \times \mathcal{A} \to \mathbb{R}$ is a reward function, and $\gamma \in [0, 1)$ is a discount factor. A stochastic policy $\pi(a| s)$ maps a given state $s$ to a probability distribution over actions $\pi: \mathcal{S} \to \mathcal{A}$. The objective of the agent is to maximize the expected cumulative discounted reward $\mathbb{E}_{\pi} [\sum_{t=0}^{\infty} \gamma^t r_t]$, where $r_t$ is the obtained reward at the discrete time step $t$.
	
	\paragraph{Two-Layer HRL Framework:}
	We adopt a continuous control RL setting, modeled as a finite-horizon, goal-conditioned MDP
	$M=<\mathcal{S}, \mathcal{G}, \mathcal{A}, \mathcal{P}, \mathcal{R}, \gamma>$,  where $\mathcal{G}$ is a goal set.  We consider a HRL framework comprising two hierarchies following \cite{NachumGLL18} with a high-level policy $\pi_h(g|s)$ and a low-level policy $\pi_l(a|s, g)$. High-level policy operates at a coarser layer and generates a high-level 
	action, \ie, subgoal $g_t \sim \pi_h (\cdot| s_t) \in \mathcal{G}$, every $k$ timesteps when $t\equiv 0~($mod $k)$.
	A pre-defined goal transition
	function $g_t=f(g_{t-1}, s_{t-1}, s_t)$ is utilized when $t\not\equiv0 ~($mod $k)$. The high level modulates the behavior of the low-level policy by intrinsic rewards for reaching these subgoals. Following prior work \cite{AndrychowiczCRS17,NachumGLL18,ZhangG0H020}, the goal set $\mathcal{G}$ corresponds to a subset of state space, \ie, $\mathcal{G} \subset \mathcal{S}$, and the goal transition function is defined as $f(g_{t-1}, s_{t-1}, s_t) = s_{t-1}+g_{t-1}-s_t$. The high-level policy aims to maximize the extrinsic reward $r_{kt}^h$ defined as:
	\begin{align}
	r_{t}^h=\sum _{i=t}^{t+k-1} r_i^{\text{env}}, t=0,1,2,\cdots
	\label{eq:1}
	\end{align}
	where $r_i^{\text{env}}$ is the reward from the environment. 
	
	The low-level policy aims to maximize the intrinsic reward provided by the high-level policy. It takes the high-level action or subgoal $g$ as input, and interacts with the environment every timestep by taking an action $a_t \sim \pi_l(\cdot| s_t, g_t) \in \mathcal{A}$. To encourage the low-level policy to reach the subgoal $g$, an intrinsic reward function measuring the subgoal-reaching performance is adopted $r_t^l = -||s_t+g_t-s_{t+1}||_2$. 
	
	The above goal-conditioned HRL framework allows the low-level policy to receive learning signals even before achieving a certain goal-reaching capability and enables concurrent end-to-end training of the high-level and low-level policies. However, off-policy training of the above HRL framework suffers from the non-stationarity problem of the high-level policy as mentioned in Section 1. HIRO \cite{NachumGLL18} proposes to relabel the high-level transition $(s_t, g_t, \sum_{i=t}^{t+k-1} r_i^{\textmd{env}}, s_{t+k} )$ with a different subgoal $\tilde{g_t}$ to make the actual observed low-level action sequence more likely to have happened with respect to the current low-level policy by maximizing $\pi_l(a_{t:t+k-1}|s_{t:t+k-1}, \tilde{g}_{t:t+k-1})$. 
	
	\section{State-Conditioned Adversarial Subgoal Generation}
	\label{method}
	
	In this section, we present our {\em State-conditioned Adversarial subGoal generAtion for hierarchical learning} (SAGA), an adversarial learning approach guiding the high-level policy
	generating more reachable subgoals for low-level policy. 
	The non-stationarity in the previous HRL methods leads to unstable and inefficient high-level policy training. In this section, we introduce our 
	adversarial learning approach to significantly improve the sample efficiency and overall performance of off-policy training of the high-level policy.

	\paragraph{Adversarial Learning of High-Level Policy:}
	SAGA integrates adversarial learning and policy training in a two-player game similarly to Generative Adversarial Networks (GANs) \cite{GoodfellowPMXWOCB14}, which primarily comprises a subgoal generator network $G(\mathbf{s}; \theta_g): \mathbf{s} \to \mathbf{g}$ and a subgoal discriminator network $D(\mathbf{g} | \mathbf{s}; \theta_d) \to \{0, 1\}$. As opposed to the generator defined in GAN which samples from a noise distribution, our subgoal generator network $G(\mathbf{s}; \theta_g)$ maps from state space to subgoal space; our subgoal discriminator network is conditioned on state $\mathbf{s}$ reminiscent of conditional GAN \cite{mirza2014conditional} where, in contrast,  both its generator and discriminator are conditioned on class labels. 
	In order to mitigate the non-stationary issue, we aim to reduce the divergence between the data distribution of the relabeled experience and the current high-level policy behaviour, with the assumption that subgoals in the relabeled experience are ``optimal'' in learning a stationary hierarchical model. To this end, the subgoal discriminator tries to distinguish the generated subgoals from relabeled subgoals of the replay buffer. In practice, we let the subgoal generator $G(\mathbf{s}; \theta_g)$ be a surrogate of the high-level actor network. 
	
	Although our approach is applicable to general actor-critic based HRL algorithms, we adopt the TD3 \cite{fujimoto2018addressing} algorithm for each level in the HRL structure following HIRO \cite{NachumGLL18} and HRAC \cite{ZhangG0H020}. Thus the first objective of the subgoal generator is to 
	maximize the expected return induced by a deterministic policy:
	\begin{align}
	J_{\textmd{dpg}} = \mathbb{E}_{\mathbf{s}  \sim \mathcal{D}} [Q_h (\mathbf{s}, \mathbf{g}) |_{\mathbf{g}=G(\mathbf{s}; \theta_g)} ]
	\label{eq:2}
	\end{align}
	where $\mathcal{D}$ is the replay buffer with the high level action relabeled similarly to HIRO, \ie, relabeling $g_t$ of the high level transition $(s_t, g_t, \sum_{i=t}^{t+k-1} r_i^{\textmd{env}}, s_{t+k} )$ with $\tilde{g_t}$ to maximize the probability of incurred low-level action sequence $\pi_l(a_{t:t+k-1}| s_{t:t+k-1}, \tilde{g}_{t:t+k-1})$, which is approximated by maximizing the log probability 	
	\begin{align}
	\textmd{log} ~\pi_l(a_{t:t+k-1}| s_{t:t+k-1}, \tilde{g}_{t:t+k-1}) \propto \nonumber\\
	-\frac{1}{2} \sum_{i=t}^{t+k-1} || a_i - \pi_l (s_i, \tilde{g_i}) ||_2^2 + \textmd{const.} 
	\label{eq:3}
	\end{align}
	
	In order to learn the distribution of the subgoal generator $G(\mathbf{s}; \theta_g)$ over the relabeled subgoal $\tilde{g}$ through adversarial learning, we define a subgoal discriminator network $D(\mathbf{g} | \mathbf{s} ; \theta_d)$ which outputs the
	probability that subgoal $g$ is an ``optimal'' subgoal, \ie, the relabeled subgoals rather than a subgoal sampled from the generator's distribution $G(\mathbf{s})$. 
	That is, we train $D(\mathbf{g} | \mathbf{s}; \theta_d)$ to maximize the probability
	of distinguishing the data distribution of ``optimal'' and ``sub-optimal'' subgoals.  Simultaneously we train $G(\mathbf{s}; \theta_g)$ to minimize the probability that
	a generated subgoal is classified as a ``sub-optimal'' subgoal by the discriminator network, that is, we minimize $\textmd{log} (1-D(G(\mathbf{s})| \mathbf{s}))$:
	\begin{align}
	J_{\textmd{adv}} = \mymin_G  \mymax_D V(D, G) = \mathbb{E}_{\mathbf{s},\mathbf{g}\sim \mathcal{D}} [\textrm{log} D(\mathbf{g} | \mathbf{s})] + \nonumber \\ \mathbb{E}_{\mathbf{s} \sim \mathcal{D}} [\textrm{log} (1-D(G(\mathbf{s} )| \mathbf{s})) ]. 
	\label{eq:4}
	\end{align}
	
	Combining terms defined in Eq. (\ref{eq:2}) and Eq. (\ref{eq:4}), the high-level actor \ie, subgoal generator $G(\mathbf{s}; \theta_g)$  is learned by performing gradient update on parameter $\theta_g$
	\begin{align}
	\bigtriangledown_{\theta_g} J = \mathbb{E}_{\mathbf{s}  \sim \mathcal{D}} [\bigtriangledown_{\theta_g} G(\mathbf{s})  \bigtriangledown_{\mathbf{g}} Q_h (\mathbf{s}, \mathbf{g}) |_{\mathbf{g}=G(\mathbf{s})} ] \nonumber \\
	- \alpha_{\textmd{adv}}  \mathbb{E}_{\mathbf{s} \sim \mathcal{D}} [\bigtriangledown_{\theta_g}  \textrm{log} (1-D(G(\mathbf{s} )| \mathbf{s})], 
	\label{eq:5}
	\end{align}
	where $\alpha_{\textmd{adv}} $ is a hyperparameter to weigh the adversarial loss.
	
	The subgoal discriminator is learned by updating $\theta_d$ with gradient
	\begin{align}
	\bigtriangledown_{\theta_d} J_{\textmd{adv}} = \mathbb{E}_{\mathbf{s}, \mathbf{g} \sim \mathcal{D}} [\bigtriangledown_{\theta_d}  [\textrm{log} D(\mathbf{g} | \mathbf{s}) + \textrm{log} (1-D(G(\mathbf{s} )| \mathbf{s})] ]. \nonumber
	\end{align}

	Note that SAGA is not enforcing the high level to generate the exact relabeled actions, but rather encourages the high-level action to follow a similar data distribution with the one which is compatible with the current instantiation of the low-level policy regardless of its effectiveness. Consequently, the approach attempts to stabilize the hierarchical learning with minimum risk of hurting exploration. 
	
	\begin{figure*}[!t]
		\includegraphics[width=0.99\linewidth]{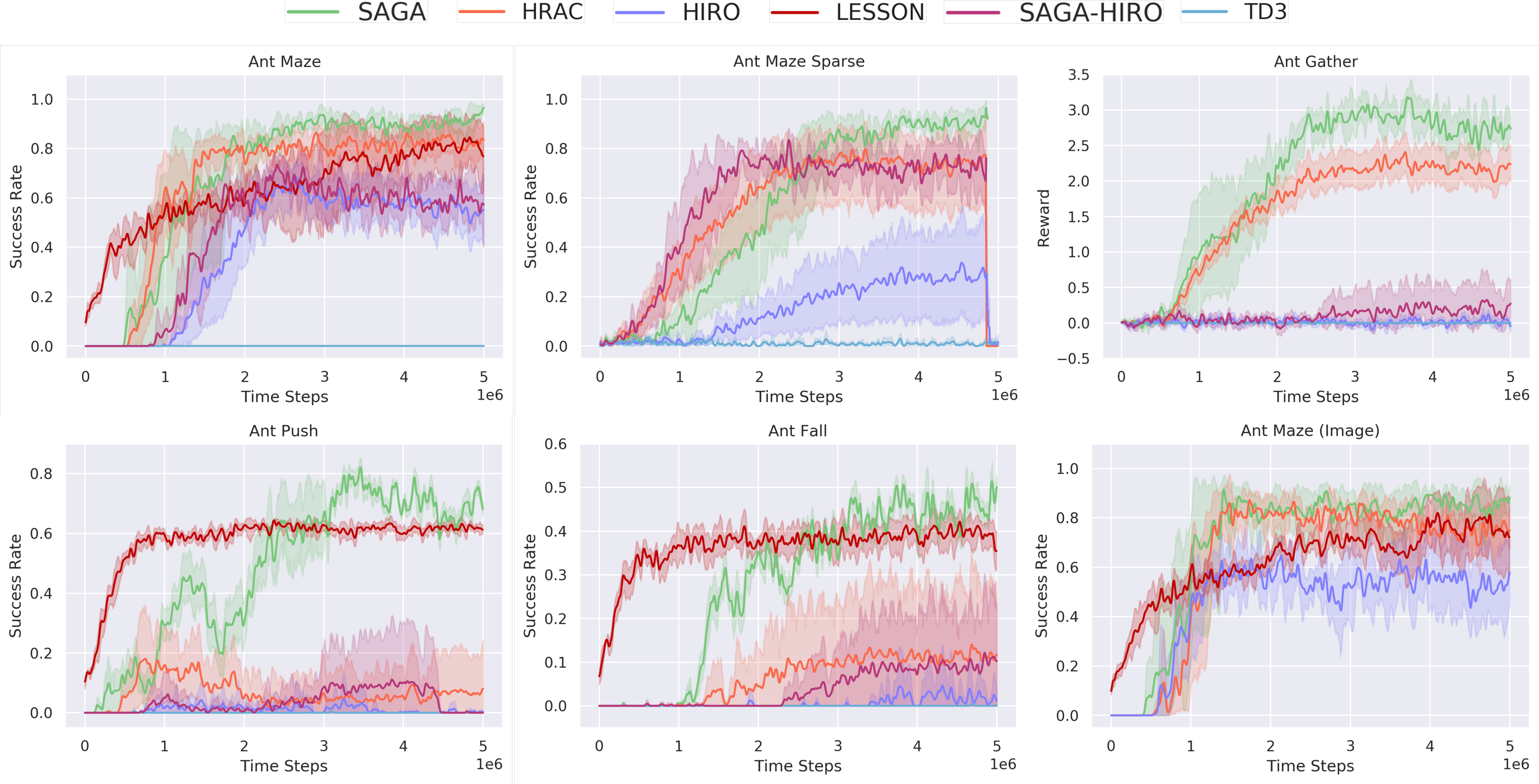}
		\centering
		\caption{Learning curves of SAGA and baselines on all environments. Each curve and its shaded region represent average episode reward (for Ant Gather) or average success rate (for the rest; see the appendix) and 95\% confidence interval respectively, averaged over 10 independent trials.  We find that SAGA performs well across all tasks. It is worth noting that SAGA learns rapidly; on the complex navigation tasks it normally requires only less than three million environment steps to achieve good performance. }
		\label{fig:comparison}
	\end{figure*}
	
	\begin{table*}[t]
		\centering
		\begin{small}
			\begin{tabular}{l|c|c|c|c|c|c}
				\hline
				& Ant Maze &   Ant Maze Sparse & Ant Gather & Ant Push & Ant Fall & Ant Maze (Image)\\
				\hline
				SAGA           & \textbf{0.93$\pm$0.01}   & \textbf{0.92$\pm$0.01} & \textbf{2.72$\pm$0.07}  &  \textbf{0.72$\pm$0.02}  & \textbf{0.47$\pm$ 0.02} & \textbf{0.87$\pm$ 0.02}\\
				HRAC           & 0.83$\pm$0.03   & 0.75$\pm$0.08 & 2.19$\pm$0.34  &  0.06$\pm$0.06  & 0.11 $\pm$0.07 & 0.76 $\pm$0.03 \\
				HIRO            & 0.54$\pm$0.06  & 0.29$\pm$0.10 & 0.02$\pm$0.02  &  0.00$\pm$0.00    & 0.01$\pm$ 0.01 & 0.53$\pm$ 0.06\\
				LESSON        & 0.81$\pm$0.04   & -                        & -                         &  0.62$\pm$ 0.02 & 0.38$\pm$ 0.01 & 0.74$\pm$ 0.06\\
				TD3              & 0.0$\pm$0.0      & 0.01$\pm$0.0    & 0.0$\pm$0.0      &  0.0$\pm$0.0     & 0.0$\pm$0.0       & -\\
				\hline
				w/o state for D           & 0.91$\pm$0.02  & 0.88$\pm$0.01 & 2.62$\pm$0.06  &  0.58$\pm$0.03  & 0.39$\pm$ 0.03 & -\\
				w/o state for D/G             & 0.0$\pm$0.0       & 0.19$\pm$0.01    & 0.0$\pm$0.0      &  0.0$\pm$0.0      & 0.0$\pm$0.0 & - \\
				SAGA-HIRO & 0.58$\pm$0.03   & 0.72$\pm$0.04 & 0.18$\pm$0.15  &  0.0$\pm$0.0      & 0.11$\pm$ 0.07 & -\\
				\hline
			\end{tabular}
		\end{small}
		\caption{Final performance of the policy obtained after 5M steps of training, averaged over 10 randomly seeded trials with standard error. }
		\label{table:quantitative}
	\end{table*}

	\section{Related Work}
	
	HRL \cite{DayanH92, schmidhuber1993planning,kulkarni2016hierarchical,vezhnevets2017feudal,NachumGLL18,LevyKPS19,ZhangG0H020,li2020learning} has long held the promise to tackle long-term credit assignment and sparse reward problems, where the high-level policy decomposes the task into subtasks whilst the low-level policy learns how to efficiently solve these subtasks. The specific way of this decomposition, \ie,~how exactly the high level communicates with the low level, varies in different approaches. Various forms of signals from the high level have been proposed, ranging from using discrete value for option \cite{bacon2017option,FoxKSG17,GregorRW17}  or skill \cite{konidaris2009efficient,EysenbachGIL19,SharmaGLKH20,bagaria2019option} selection, to forming a vector within a learned embedding space as subgoal \cite{vezhnevets2017feudal,li2020learning}.  However, majority of these approaches are unable to benefit from advances of off-policy model-free RL.  
	
	Improving the learning efficiency of HRL through off-policy training has attracted a considerable amount of research efforts in recent years. However, besides instability, off-policy training also poses the non-stationary problem which is characteristic to
	HRL. 
	In \cite{NachumGLL18}, they proposed an off-policy method which relabels past experiences to reduce the impact in training with
	invalid high-level state transitions due to non-stationarity. Employing hindsight techniques \cite{AndrychowiczCRS17}, \cite{LevyKPS19} proposed to train multi-level policies in parallel while penalizing the high-level for generating subgoals which are not reachable in the low level. In \cite{ZhangG0H020} the large subgoal space issue was addressed by restricting the high-level action space from the whole subgoal space using an adjacency constraint. In \cite{wang2020i2hrl}  high-level policy decision making is conditioned on the received low-level policy representation as well as the state of the environment to improve stationarity. 
	Another solution is to add a slowness objective to effectively learn the subgoal representation so that the low-level reward function varies in a stationary way~\cite{li2020learning}. 
	
	The general topic of goal generation in RL has also been studied \cite{FlorensaHGA18,NairPDBLL18,RenD00019,CamperoRKTRG21}. GoalGAN~\cite{FlorensaHGA18} uses a standard GAN to produce tasks at the appropriate level of difficulty for training the policy. While GoalGAN is similar in spirit with the proposed SAGA to some extent, there are several key differences apart from if it is a hierarchical policy or not. GoalGAN is using a standalone generator that does not condition on the observation; its GAN and policy are two modules that are independently and sequentially trained. In contrast, SAGA's generator is a surrogate of the original actor network and SAGA directly updates its policy through the incurred adversarial loss and policy loss concurrently. Other related works are, for example, \cite{NairPDBLL18} that proposes to combine unsupervised representation learning and reinforcement learning of goal-conditioned policies. A framework to generate  hindsight goals which are easy for an agent to achieve in the short term is proposed in \cite{RenD00019}. In a recent work \cite{CamperoRKTRG21} a framework where a teacher network learns to propose increasingly challenging yet achievable goals is proposed; the teacher is positively rewarded if the student achieves the goal with suitable effort, but penalized if the student either cannot achieve the goal, or can do so too easily. The foremost difference from SAGA is that these methods are developed for flat architectures and therefore cannot successfully solve tasks which require complex high-level decision making. 
	
	\section{Experiments}
	
	This section evaluates and compares our method against standard RL and prior HRL methods in challenging environments which require a combination of locomotion and object manipulation. We also ablate the various components to understand their importance. Our experiments are designed to answer the following questions:
	
	\begin{enumerate} 
		\item{} Can SAGA improve the sample efficiency and performance of goal-conditioned HRL across various long-horizon continuous control tasks?  
		\item{} Can SAGA outperform alternative adversarial learning approaches in the goal-conditioned HRL framework? 
	\end{enumerate} 
	
	\subsection{Environment Setup}
	We consider the following five environments for our analysis: 
	\begin{enumerate} 
		\item{} \textbf{Ant Maze}: A `$\supset$'-shaped maze poses a challenging navigation task for a quadruped-Ant.  The ant needs to reach a target position starting from a random position in a maze with dense rewards. A variant (labeled `Image') with low-resolution image observations is also adopted;  the observation is formed by zeroing out the x, y coordinates and appending a 5$\times$5$\times$3 top-down view of the environment, as described in \cite{NachumGLL19}.
		\item{} \textbf{Ant Maze Sparse}: From a random start position, the ant needs to reach a target position in a maze with sparse rewards.
		\item{} \textbf{Ant Gather}: Starting from a fixed position, the ant collects green apples and avoids red bombs. 
		\item{} \textbf{Ant Push}: A challenging environment which requires both task and motion planning. The ant needs to move to the left of the maze so that it can move up and right to push the block out of the way for reaching the target. 
		\item{} \textbf{Ant Fall}: This environment extends the navigation to three dimensions. The ant starts on a raised platform with the target located directly in front of it but separated by a chasm which it cannot cross by itself. The ant needs to push the block forward, fill the gap, walk across and move to the left in order to reach the target. 
	\end{enumerate} 

	\begin{figure*}[t]
		\includegraphics[width=0.99\linewidth]{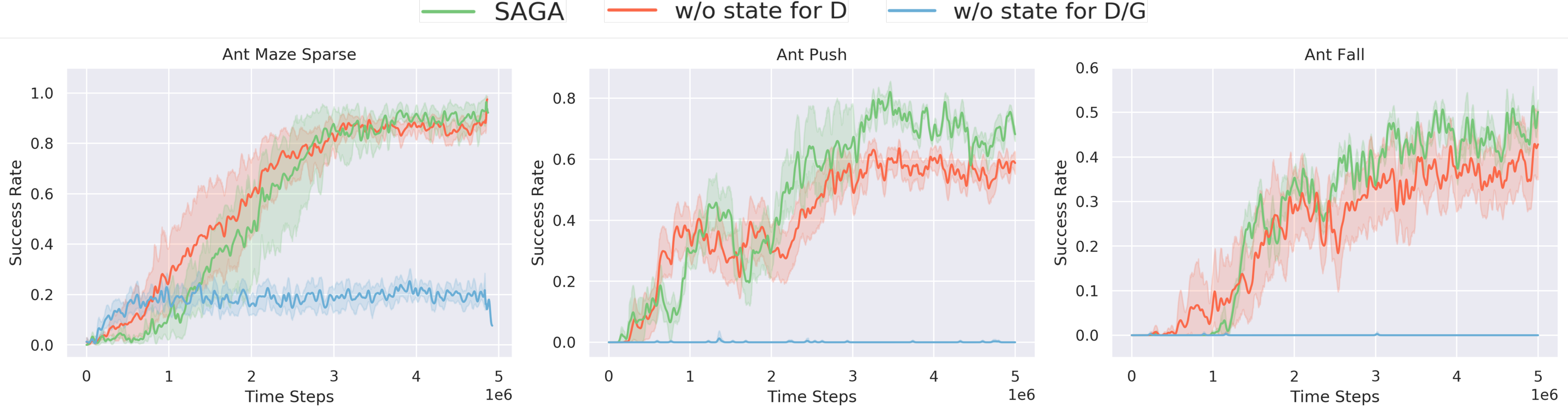}
		\centering
		\caption{Ablation studies of adversarial learning approaches, averaged over 10 independent trials.}
		\label{fig:ablation}
	\end{figure*}

	\subsection{Implementation}
	
	For the hierarchical policy network, we employ the same architecture as HRAC \cite{ZhangG0H020} which adopts TD3 \cite{fujimoto2018addressing} as the underlying algorithm for training both the high-level and low-level policy.  Specifically,
	we adopt two networks comprising three fully-connected layers with ReLU nonlinearities as the actor and critic networks
	of both low-level and high-level TD3 networks. The size of the hidden layers of both actor and critic is 300. The output
	of the high-level actor is activated using the tanh function and scaled according to the size of the environments. 
	
	The subgoal generator network has the identical architecture as the high-level actor. For the subgoal discriminator network, we use
	a network consisting of 3 fully-connected layers (size of 300, 300 and 1 respectively) with Leaky-ReLU (negative slope 0.2) nonlinearities and sigmoid function in all tasks. Adam optimizer is used for all networks. 
	

	\subsection{Comparative Analysis}
	
	To test the performance of SAGA, we compare against the following baseline methods:
	\begin{enumerate}
		\item \textbf{HIRO} \cite{NachumGLL18}: an off-policy goal-conditioned HRL algorithm proposes to address the non-stationarity issue by relabeling high-level actions. 
		\item\textbf{HRAC} \cite{ZhangG0H020}: a state-of-the-art off-policy goal-conditioned HRL algorithm introduces an adjacency network to restrict the high-level action space to a $k$-step adjacent region of the current state\footnote{We use HRAC's official implementation \url{https://github.com/trzhang0116/HRAC} to evaluate both HRAC and HIRO since HRAC is built on HIRO which provides fair comparisons using the same HRL implementation. }
		\item\textbf{LESSON} \cite{li2020learning}: a state-of-the-art off-policy goal-conditioned HRL algorithm learns the subgoal representation by posing a slowness objective. 
		\item\textbf{TD3}  \cite{fujimoto2018addressing}: a state-of-the-art flat RL algorithm we compare to validate the need for hierarchical policies.
	\end{enumerate}

	\begin{figure*}[t]
		\includegraphics[width=0.99\linewidth]{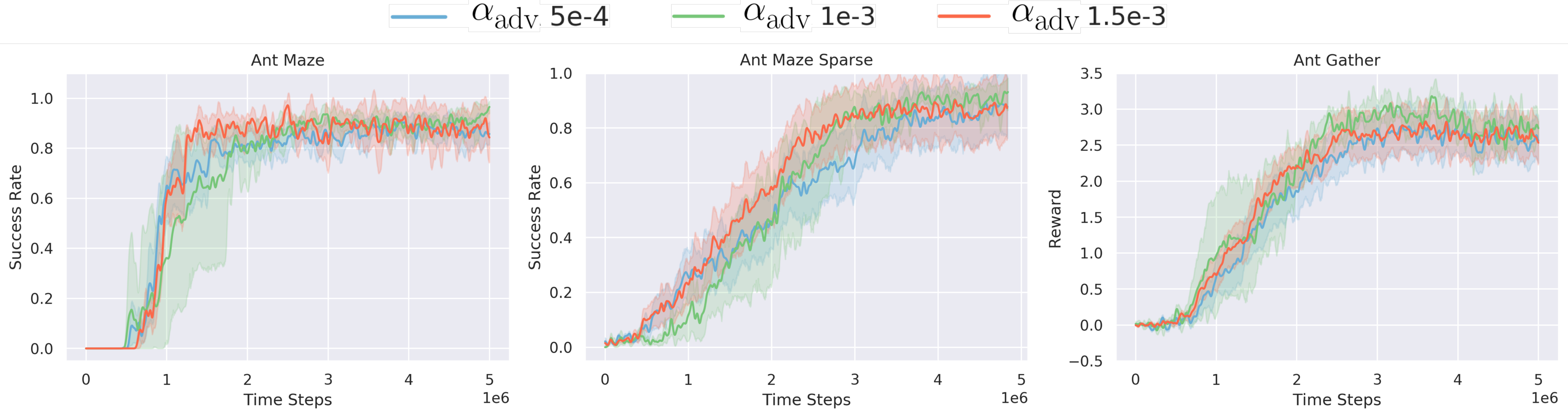}
		\centering
		\caption{Learning curves with different coefficient of adversarial loss $\alpha_{\textmd{adv}}$, averaged over 5 independent trials.}
		\label{fig:weight}
	\end{figure*}

	For fair comparison,  all the HRL baselines use the same hierarchical structure and environment configuration as SAGA.
	Table \ref{table:quantitative} shows the final performance of the policy after training, and all baselines are significantly outperformed by SAGA. 
	The learning curves of SAGA and baselines across all tasks are plotted in Fig. \ref{fig:comparison}. In the gather task \ie, Ant Gather, SAGA achieves 
	considerably better performance and consistently exceeds all baselines in gather task  \ie, Ant Gather, and all navigation tasks \ie, Ant Maze, Ant Maze (Image), Ant Maze Sparse, Ant Push and Ant Fall\footnote{We use LESSON's official implementation \url{https://github.com/SiyuanLee/LESSON} which includes environments Ant Maze, Ant Push and Ant Fall and adapt its task settings to be the same with other baselines for fair comparisons}, in terms of sample efficiency and asymptotic performance. SAGA shows consistently better training efficiency benefiting from the improved learning stability of the hierarchical policies. This suggests that the proposed adversarial learning approach effectively enforces the high-level policy to generate subgoals compatible with the current instantiation of the low-level policy during training which in turn significantly mitigates the non-stationarity issue of off-policy training in HRL. It is also observed that flat RL algorithm TD3 does not learn in the complex environments used in the experiments which further validate the need for hierarchical policies. 
	
	\subsection{Ablative Analysis}
	
	We also compare SAGA with several variants to investigate the importance of various design choices of  SAGA. As we employ HRAC as the base model for SAGA, we introduce a variant of SAGA which is employing HIRO as base model to understand the generalization of our proposed approach. Fig. \ref{fig:comparison} also shows the learning curves of SAGA-HIRO and HIRO. In the Ant Maze task with dense rewards, SAGA-HIRO achieves slightly better performance with HIRO, while SAGA-HIRO exceeds HIRO in other tasks. Table \ref{table:quantitative} shows the consistent quantitative improvements by introducing the proposed approach. We note that HIRO hardly learns in Ant Push and learns poorly in Ant Fall by using the standard RL training based on the relabeled high-level actions, whereas SAGA-HIRO enforces the policy to learn in a more sample efficient manner. The empirical evaluation confirms that our algorithm is a principled and generic approach which can be applied in existing goal-conditioned HRL methods to effectively address the non-stationarity issue.
	
	In the ablation studies of alternative adversarial learning approaches in the goal-conditioned HRL framework, we introduce two variants: 
	\begin{enumerate}
		\item\textbf{w/o state for D}: a variant that uses a discriminator network which is not conditioned on state;
		\item\textbf{w/o state for D/G}: a variant that uses common generator and discriminator networks which neither are conditioned on state, \ie,
		the generator network takes as input a random noise sampled from Normal distribution and then trains the high-level critic using the generated subgoals, similar to \cite{FlorensaHGA18} in flat RL case. 
	\end{enumerate}
	As shown in Table \ref{table:quantitative} and Fig. \ref{fig:ablation},  the advantage of the discriminator network conditioning on states is more pronounced in challenging tasks Ant Push and Ant Fall, where subgoal distributions may heavily depend on the states. In other words, state-conditioned discriminator network is able to account for the dynamic elements in the environment and subsequently enables subgoals that specifically interact with those elements. In the second variant, \ie, a vanilla GAN setting, the generator network is no longer a surrogate of the high-level actor network. The assumption that both generator and discriminator are not depending on states leads to slow learning of generator network and consequently the deteriorating non-stationarity issue for the hierarchical policies. 
	
	\subsection{Analysis of Hyperparameter Selection}
	
	We empirically study the effect of different coefficients of adversarial loss $\alpha_{\textmd{adv}}$. Fig. \ref{fig:weight} shows that SAGA with three coefficients of adversarial loss 0.0005, 0.001 and 0.0015 shows asymptotically similar results and  generally $\alpha_{\textmd{adv}}=0.001$ gives better performance across three tasks; we use  $\alpha_{\textmd{adv}}=0.001$  for all the tasks presented in the paper. In general, larger $\alpha_{\textmd{adv}}$ implies that the learning prioritizes the adaptation of the generated subgoals to follow the distribution of relabeled subgoals which speeds up its learning process harnessing the more accurate learning signals provided. However, the learning process may slow down in case of very large $\alpha_{\textmd{adv}}$ since the learning of the high-level value function will be slowed down and its subsequent exploration might be affected. As a contrary, with smaller $\alpha_{\textmd{adv}}$  the standard RL training is more pronounced; SAGA eventually degenerates to original baseline when its value is considerably small. 
	
	\begin{table*}[!t]
		\centering
		\begin{small}
			\begin{tabular}{l|c|c|c|c|c}
				\hline
				& Ant Maze &   Ant Maze Sparse & Ant Gather & Ant Push & Ant Fall \\
				\hline
				SAGA & \textbf{2.79$\pm$0.08} & \textbf{2.13$\pm$0.26}  & \textbf{1.65$\pm$0.52} & \textbf{2.62$\pm$0.46} & \textbf{2.49$\pm$ 0.23} \\
				HRAC  & 3.41$\pm$0.31  & 4.38$\pm$0.84  & 3.59$\pm$0.91  & 4.95$\pm$1.11 & 4.10 $\pm$1.03\\
				HIRO  & 10.74$\pm$1.05 & 14.14$\pm$0.0 & 13.71$\pm$1.15  & 9.38$\pm$2.85 & 11.4$\pm$ 1.69\\
				\hline
			\end{tabular}
		\end{small}
		\caption{The distance between generated subgoals and the reached subgoals \ie, the final state of k-step low-level roll-out, averaged over 10 randomly seeded trials with standard error. }
		\label{table:distance}
	\end{table*}
	
	\begin{figure*}[!t]
		\centering
		\includegraphics[width=0.99\linewidth]{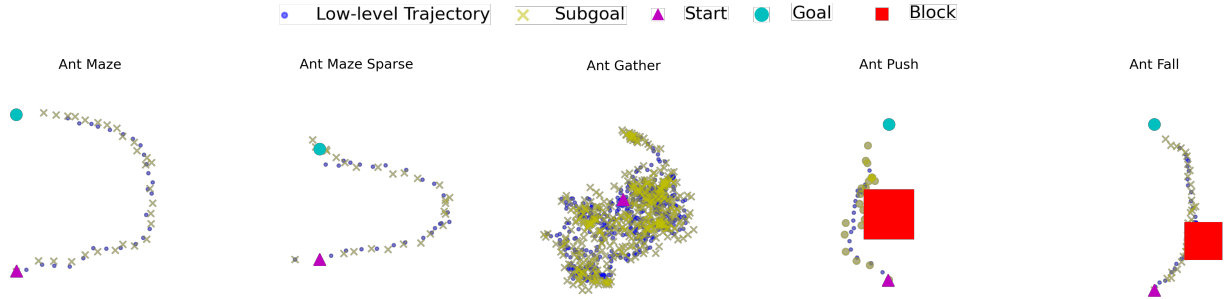}
		\caption{Visualization of generated subgoals by SAGA and the reached subgoals \ie, the final state of k-step low-level roll-out, in one of the randomly seeded trials. The generated subgoals are reachable and generally match the low-level trajectories. }
		\label{fig:traj}
	\end{figure*}
	
	\begin{figure*}[!t]
		\centering
		\includegraphics[width=0.99\linewidth]{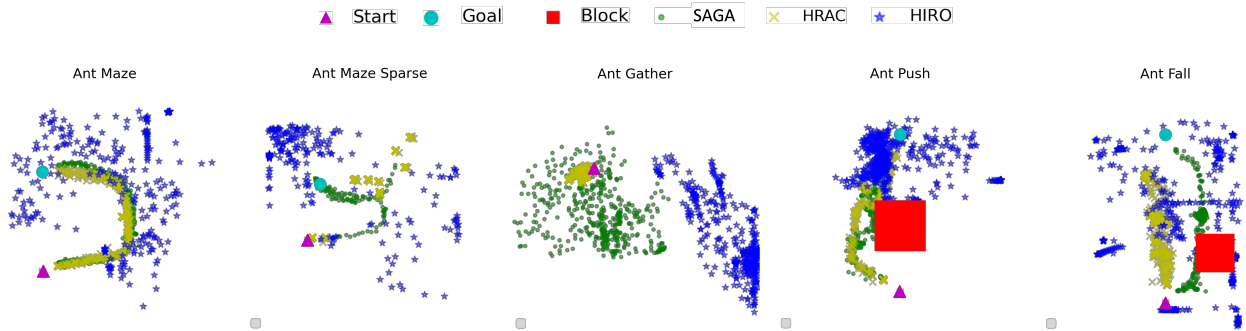}
		\caption{Visualization of generated subgoals. Subgoals generated by SAGA are generally matching the low-level trajectories or planed motions, which indicates that SAGA can generate reasonable subgoals for the low level to achieve and also guide the optimization to jump out of the local optimum of the ant to move directly towards the target in complex tasks such as Ant Push and Ant Fall. In contrast, subgoals generated by HRAC and HIRO frequently get stuck to local minimum and fail to guide the agent to accomplish the final task. Subgoals generated by LESSON lie in a learned subgoal representation space and may not be visualized along with other methods. }
		\label{fig:subgoals}
	\end{figure*}
	
	\subsection{Analysis of Generated Subgoals}
	We visualize the generated subgoals of SAGA and the reached subgoals \ie, the final state of k-step low-level roll-out in Fig.~\ref{fig:traj}. The subgoals generated by SAGA are generally reachable and match the low-level trajectories, since they are distributed in the close vicinity of the actually reached subgoals. This is further confirmed by the measure of distance between generated subgoals and the reached subgoals \ie, the final state of k-step low-level roll-out in Table \ref{table:distance}. This suggests that SAGA generates the most reachable subgoals and may give the most effective learning signal for the low level to improve sample efficiency compared with HRAC and HIRO. 
	
	We also visually compare the generated subgoals of SAGA, HIRO and HRAC in Fig.~\ref{fig:subgoals}. The subgoals generated by SAGA generally match the planned motions in Ant Maze, Ant Maze Sparse and Ant Gather. Notably, SAGA generates subgoals to guide the optimization to jump out of the local optimum of the ant to move directly towards the target in Ant Push and Ant Fall. In detail, under the hierarchical policy trained by SAGA, in Ant Push the ant first moves to the left, then pushes the block to the right and finally reaches the target; guided by subgoals generated by SAGA, in Ant Fall, the ant first moves to the right to push the block forward which fills the gap and then walks across and moves to the left in order to finally reach the target. HRAC can also generate relatively reasonable subgoals to reach based on affinity constraints for the low level, however these subgoals frequently get stuck in a local optimum of moving directly to the target as illustrated in Ant Push and Ant Fall.  HIRO, on the contrary, fails to generate achievable subgoals and cannot guide the agent to achieve its final target.

	\section{Conclusion}
	
	We proposed a novel adversarially guided subgoal generation framework for goal-conditioned HRL to mitigate the issue of non-stationarity in off-policy training. The learning of high-level policy is formulated as a two-player game where the subgoal generator endeavours to generate subgoals compatible with the current instantiation of low-level policy while the proposed discriminator network tries to distinguish the generated subgoals from the relabled subgoals. Empirical studies show that the proposed adversarial learning is capable of reducing the shifts in data distribution from relabeled experience to the current high-level policy behaviour and consequently improving the overall learning efficiency and stability.

	\bibliography{aaai23} 
	
\end{document}